\documentclass[11pt,letterpaper]{article}

\usepackage{emnlp2016}
\usepackage{pslatex}
\usepackage[english]{babel}
\usepackage[utf8]{inputenc}
\usepackage{amsmath}
\usepackage{bm}
\usepackage{graphicx}
\usepackage{tikz}
\usepackage{xcolor}
\usepackage{url}
\usepackage{rotating}
\usepackage{natbib}

\newcommand{\soft}[1]{}
\newcommand{\nopreview}[1]{}

\emnlpfinalcopy
\def\emnlppaperid{496}

\title{Modeling Human Reading with Neural Attention}

\author{Michael Hahn \quad  Frank Keller\\
Institute for Language, Cognition and Computation\\
School of Informatics, University of Edinburgh \\
10 Crichton Street, Edinburgh EH8 9AB, UK\\
\url{s1582047@inf.ed.ac.uk} \quad \url{keller@inf.ed.ac.uk}
}

\date{}

\begin{document}

\maketitle

\begin{abstract}
  When humans read text, they fixate some words and skip others.
  However, there have been few attempts to explain skipping behavior
  with computational models, as most existing work has focused on
  predicting reading times (e.g.,~using surprisal). In this paper, we
  propose a novel approach that models both skipping and reading,
  using an unsupervised architecture that combines a neural attention
  with autoencoding, trained on raw text using reinforcement learning.
  Our model explains human reading behavior as a tradeoff between
  precision of language understanding (encoding the input accurately)
  and economy of attention (fixating as few words as possible). We
  evaluate the model on the Dundee eye-tracking corpus, showing that
  it accurately predicts skipping behavior and reading times, is
  competitive with surprisal, and captures known qualitative features
  of human reading.
\end{abstract}

\section{Introduction}
\label{sec:intro}

Humans read text by making a sequence of fixations and saccades.
During a fixation, the eyes land on a word and remain fairly static
for 200--250~ms. Saccades are the rapid jumps that occur between
fixations, typically lasting 20--40~ms and spanning 7--9~characters
\citep{rayner_eye_1998}. Readers, however, do not simply fixate one
word after another; some saccades go in reverse direction, and some
words are fixated more than once or skipped altogether.

A range of computational models have been developed to account for
human eye-movements in reading \citep{rayner_models_2010-1}, including
models of saccade generation in cognitive psychology, such as
EZ-Reader \citep{reichle_toward_1998, reichle_ez_2003,
  reichle_using_2009}, SWIFT \citep{engbert_dynamical_2002,
  engbert_swift:_2005}, or the Bayesian Model of
\citet{bicknell_rational_2010-1}. More recent approaches use machine
learning models trained on eye-tracking data to predict human reading
patterns \citep{nilsson_learning_2009, nilsson_towards_2010,
  hara_predicting_2012, matthies_blinkers_2013}. Both types of models
involve theoretical assumptions about human eye-movements, or at least
require the selection of relevant eye-movement features. Model
parameters have to be estimated in a supervised way from eye-tracking
corpora.

Unsupervised approaches, that do not involve training the model on
eye-tracking data, have also been proposed. A key example is
\emph{surprisal}, which measures the predictability of a word in
context, defined as the negative logarithm of the conditional probability of
the current word given the preceding words
\citep{hale_probabilistic_2001, levy_expectation-based_2008}.
Surprisal is computed by a language model, which can take the form of
a probabilistic grammar, an n-gram model, or a recurrent neural
network. While surprisal has been shown to correlate with word-by-word
reading times \citep{mcdonald_eye_2003, mcdonald_low-level_2003,
  demberg_data_2008, frank:bod:11, smith_effect_2013}, it cannot
explain other aspects of human reading, such as reverse saccades,
re-fixations, or skipping. Skipping is a particularly intriguing
phenomenon: about 40\% of all words are skipped (in the Dundee corpus,
see below), without apparent detriment to text understanding.

In this paper, we propose a novel model architecture that is able to
explain which words are skipped and which ones are fixated, while also
predicting reading times for fixated words. Our approach is completely
unsupervised and requires only unlabeled text for training.

Compared to language as a whole, reading is a recent innovation in
evolutionary terms, and people learning to read do not have access to
competent readers' eye-movement patterns as training data.  This
suggests that human eye-movement patterns emerge from general
principles of language processing that are independent of reading.
Our starting point is the \emph{Tradeoff Hypothesis}: Human reading
optimizes a tradeoff between \emph{precision} of language
understanding (encoding the input accurately) and \emph{economy} of
attention (fixating as few words as possible). Based on the Tradeoff
Hypothesis, we expect that humans only fixate words to the extent
necessary for language understanding, while skipping words whose
contribution to the overall meaning can be inferred from context.

In order to test these assumptions, this paper investigates the
following questions:
\begin{enumerate}

\item Can the Tradeoff Hypothesis be implemented in an unsupervised
  model that predicts skipping and reading times in
  \emph{quantitative} terms? In particular, can we compute surprisal
  based only on the words that are actually fixated?

\item Can the Tradeoff Hypothesis explain known \emph{qualitative}
  features of human fixation patterns? These include dependence on
  word frequency, word length, predictability in context, a contrast
  between content and function words, and the statistical dependence
  of the current fixation on previous fixations.


\end{enumerate}
To investigate these questions, we develop a generic architecture that
combines neural language modeling with recent ideas on integrating
recurrent neural networks with mechanisms of attention, which have
shown promise both in NLP and in computer vision. We train our model
end-to-end on a large text corpus to optimize a tradeoff between
minimizing input reconstruction error and minimizing the number of
words fixated. We evaluate the model's reading behavior against a
corpus of human eye-tracking data. Apart from the unlabeled training
corpus and the generic architecture, no further assumptions about
language structure are made -- in particular, no lexicon or grammar or
otherwise labeled data is required.

Our unsupervised model is able to predict human skips and fixations
with an accuracy of $63.7\%$. This compares to a baseline of $52.6\%$
and a supervised accuracy of $69.9\%$. For fixated words, the model
significantly predicts human reading times in a linear mixed effects
analysis. The performance of our model is comparable to surprisal,
even though it only fixates 60.4\% of all input words. Furthermore, we
show that known qualitative features of human fixation sequences
emerge in our model without additional assumptions.

\section{Related Work}
\label{sec:related}

%

A range of attention-based neural network architectures have recently
been proposed in the literature, showing promise in both NLP and
computer vision \citep[e.g.,][]{mnih_recurrent_2014,
  bahdanau_neural_2015}. Such architectures incorporate a mechanism
that allows the network to dynamically focus on a restricted part of
the input. Attention is also a central concept in cognitive science,
where it denotes the focus of cognitive processing. In both
language processing and visual processing, attention is known to be
limited to a restricted area of the visual field, and shifts rapidly
through eye-movements \citep{Henderson:03}.

Attention-based neural architectures either employ \emph{soft
  attention} or \emph{hard attention}.  Soft attention distributes
real-valued attention values over the input, making end-to-end
training with gradient descent possible.  Hard attention mechanisms
make discrete choices about which parts of the input to focus on, and
can be trained with reinforcement learning
\citep{mnih_recurrent_2014}. In NLP, soft attention can mitigate the
difficulty of compressing long sequences into fixed-dimensional
vectors, with applications in machine translation
\citep{bahdanau_neural_2015} and question answering
\citep{hermann_teaching_2015}. In computer vision, both types of
attention can be used for selecting regions in an image
\citep{ba_learning_2015,xu_show_2015}.


\section{The NEAT Reading Model}
\label{sec:model}

The point of departure for our model is the Tradeoff Hypothesis (see
Section~\ref{sec:intro}): Reading optimizes a tradeoff between
precision of language understanding and economy of attention.  We make
this idea explicit by proposing NEAT (NEural Attention Tradeoff), a
model that reads text and attempts to reconstruct it afterwards.
While reading, the network chooses which words to process and which
words to skip.  The Tradeoff Hypothesis is formalized using a training
objective that combines accuracy of reconstruction with economy of
attention, encouraging the network to only look at words to the extent
that is necessary for reconstructing the sentence.

\subsection{Architecture}

We use a neural sequence-to-sequence architecture
\citep{sutskever_sequence_2014} with a hard attention mechanism.
We illustrate the model in Figure~\ref{fig:architecture}, operating on
a three-word sequence $\boldsymbol w$.  The most basic components are
the \emph{reader}, labeled $R$, and the \emph{decoder}.  Both of them
are recurrent neural networks with Long Short-Term Memory (LSTM,
\citealp{hochreiter_long_1997}) units.  The recurrent reader network
is expanded into time steps $R_0, \dots, R_3$ in the figure.  It goes
over the input sequence, reading one word $w_i$ at a time, and
converts the word sequence into a sequence of vectors $h_0, \dots,
h_3$.  Each vector $h_i$ acts as a fixed-dimensionality encoding of
the word sequence $w_1, \dots, w_i$ that has been read so far.  The
last vector $h_3$ (more generally $h_N$ for sequence length $N$),
which encodes the entire input sequence, is then fed into the input
layer of the decoder network, which attempts to reconstruct the input
sequence $\boldsymbol w$. It is also realized as a recurrent neural
network, collapsed into a single box in the figure.  It models a
probability distribution over word sequences, outputting a probability
distribution $P_{Decoder}(w_i|\boldsymbol{w}_{1,\dots,i-1}, h_N)$ over
the vocabulary in the $i$-th step, as is common in neural language
modeling \citep{mikolov_recurrent_2010}. As the decoder has access to
the vector representation created by the reader network, it ideally is
able to assign the highest probability to the word sequence
$\boldsymbol{w}$ that was actually read.  Up to this point, the model
is a standard sequence-to-sequence architecture reconstructing the
input sequence, that is, performing autoencoding.


As a basic model of human processing, NEAT contains two further
components.  First, experimental evidence shows that during reading,
humans constantly make predictions about the upcoming input (e.g.,
\citealp{van_gompel_syntactic_2007}).  As a model of this behavior,
the reader network at each time step outputs a probability
distribution $P_{R}$ over the lexicon. This distribution describes
which words are likely to come next (i.e.,~the reader network performs
language modeling). Unlike the modeling performed by the decoder, $P_R$,
via its recurrent connections, has access to the previous context only.

Second, we model skipping by stipulating that only some of the input
words $w_i$ are fed into the reader network $R$, while $R$ receives a
special vector representation, containing no information about the input word,
in other cases. These are the
words that are skipped. In NEAT, at each time step during reading, the
\emph{attention module} $A$ decides whether the next word is shown to
the reader network or not. When humans skip a word, they are able to
identify it using \emph{parafoveal preview} \citep{rayner_eye_2009}.
Thus, we can assume that the choice of which words to skip takes into
account not only the prior context but also a preview of the word
itself. We therefore allow the attention module to take the input word
into account when making its decision. In addition, the attention
module has access to the previous state $h_{i-1}$ of the reader network,
which summarizes what has been read so far.  To allow for interaction
between skipping and prediction, we also give the attention module
access to the probability of the input word according to the
prediction $P_R$ made at the last time step.  If we write the decision made
by $A$ as $\omega_i \in \{0, 1\}$, where $\omega_i = 1$ means that
word $w_i$ is shown to the reader and $0$ means that it is not, we can
write the probability of showing word $w_i$ as:
\begin{equation}\label{eq:att-prob}
\begin{split}
P(\omega_{i}=1|\boldsymbol\omega_{1\dots i-1}, \boldsymbol w) \\
= P_A(w_i, h_{i-1}, P_{R}(w_i|\boldsymbol w_{1\dots i-1}, \boldsymbol\omega_{1\dots i-1}))
\end{split}
\end{equation}
We implement $A$ as a feed-forward network, followed by taking a
binary sample $\omega_i$.

We obtain the \emph{surprisal} of an input word by taking the negative
logarithm of the conditional probability of this word given the
context words that precede it:
\begin{equation}\label{eq:surp}
\begin{split}
\operatorname{Surp}(w_i|\boldsymbol w_{1\dots i-1}) = -\log P_{R}(w_i|\boldsymbol w_{1\dots i-1}, \boldsymbol\omega_{1\dots i-1})
\end{split}
\end{equation}
As a consequence of skipping, not all input words are accessible to
the reader network.  Therefore, the probability and surprisal
estimates it computes crucially only take into account the words that
have actually been fixated. We will refer to this quantity as the
\emph{restricted surprisal}, as opposed to \emph{full surprisal},
which is computed based on all prior context words.

The key quantities for predicting human reading are the fixation
probabilities in equation~(\ref{eq:att-prob}), which model fixations
and skips, and restricted surprisal in equation~(\ref{eq:surp}), which
models the reading times of the words that are fixated.



\def\la{-7.7}
\def\lb{-4.0}
\def\lc{-0.3}
\def\ld{3.4}
\def\le{5.3}
\def\laa{-5.5}
\def\lba{-2.0}
\def\lca{1.7}

\begin{figure*}[tb]
\centering
\begin{tikzpicture}[%
  block/.style = {draw, fill=blue!30, align=center, anchor=west,
              minimum height=0.65cm, inner sep=0},
  ball/.style = {circle, draw, align=center, anchor=north, inner sep=0}]


\node[rectangle,draw=none,text width=0.3cm,anchor=base] (W1) at (\laa,-1) {$w_1$};
\node[rectangle,draw=none,text width=0.3cm,anchor=base] (W2) at (\lba,-1) {$w_2$};
\node[rectangle,draw=none,text width=0.3cm,anchor=base] (W3) at (\lca,-1) {$w_3$};

\node[rectangle,draw,text width=0.3cm,anchor=base] (A1) at (\lb,-1) {A};
\node[rectangle,draw,text width=0.3cm,anchor=base] (A2) at (\lc,-1) {A};
\node[rectangle,draw,text width=0.3cm,anchor=base] (A3) at (\ld,-1) {A};

\node[rectangle,draw,text width=0.3cm,anchor=base] (R0) at (\la,-3) {$R_0$};
\node[rectangle,draw,text width=0.3cm,anchor=base] (R1) at (\lb,-3) {$R_1$};
\node[rectangle,draw,text width=0.3cm,anchor=base] (R2) at (\lc,-3) {$R_2$};
\node[rectangle,draw,text width=0.3cm,anchor=base] (R3) at (\ld,-3) {$R_3$};


\node[rectangle,draw,anchor=base] (Task) at (\le,-3) {Decoder};

\draw[->] (R0.east) to [out=0,in=180] (R1.west);
\draw[->] (R1.east) to [out=0,in=180] (R2.west);
\draw[->] (R2.east) to [out=0,in=180] (R3.west);
\draw[->] (R3.east) to [out=0,in=180] (Task.182);


\draw[->] (R0.east) to [out=10,in=250] node[pos=0.65]{$\ \ \ \ \ \ \ h_0$} (A1.225);
\draw[->] (R1.east) to [out=10,in=250]  node[pos=0.65]{$\ \ \ \ \ \ \ h_1$} (A2.225);
\draw[->] (R2.east) to [out=10,in=250]  node[pos=0.65]{$\ \ \ \ \ \ \ h_2$} (A3.225);

\draw[->] (W1.10) to [out=0,in=180] (A1.west);
\draw[->] (W2.10) to [out=0,in=180] (A2.west);
\draw[->] (W3.10) to [out=0,in=180] (A3.west);



\draw[->] (A1.south) to [out=270,in=90] (R1.north);
\draw[->] (A2.south) to [out=270,in=90] (R2.north);
\draw[->] (A3.south) to [out=270,in=90] (R3.north);


\node[rectangle,draw=none,text width=0.3cm,anchor=base] (S1) at (\laa,-1.7) {$P_{R_1}$};
\node[rectangle,draw=none,text width=0.3cm,anchor=base] (S2) at (\lba,-1.7) {$P_{R_2}$};
\node[rectangle,draw=none,text width=0.3cm,anchor=base] (S3) at (\lca,-1.7) {$P_{R_3}$};


\draw[->] (R0.east) to [out=2,in=200] (S1.west);
\draw[->] (R1.east) to [out=2,in=200] (S2.west);
\draw[->] (R2.east) to [out=2,in=200] (S3.west);

\draw[->] (S1.east) to [out=0,in=180] (A1.west);
\draw[->] (S2.east) to [out=0,in=180] (A2.west);
\draw[->] (S3.east) to [out=0,in=180] (A3.west);

\end{tikzpicture}
\caption{The architecture of the proposed model, reading a three-word
  input sequence $w_1, w_2, w_3$.  $R$ is the reader network and $P_R$
  the probability distribution it computes in each time step. $A$ is
  the attention network. At each time step, the input, its probability
  according to $P_R$, and the previous state $h_{i-1}$ of $R$ are fed
  into $A$, which then decides whether the word is read or skipped.}
  \label{fig:architecture}
\end{figure*}
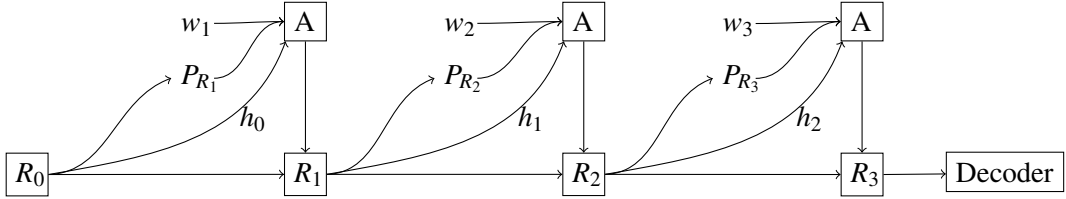

\subsection{Model Objective}


Given network parameters $\theta$ and a sequence $\boldsymbol w$ of
words, the network stochastically chooses a sequence
$\boldsymbol\omega$ according to~(\ref{eq:att-prob}) and incurs a loss
$L(\boldsymbol \omega|\boldsymbol w,\theta)$ for language modeling and
reconstruction:
\begin{equation}
\begin{split}
  L(\boldsymbol\omega|\boldsymbol w, \theta) = - \sum_i \log
  P_R(w_i|\boldsymbol{w}_{1,\dots,i-1}, \boldsymbol\omega_{1,\dots,i-1}; \theta) \\
  - \sum_i \log P_\mathit{Decoder}(w_i|\boldsymbol{w}_{1,\dots,i-1};  h_N; \theta)
\end{split}
\end{equation}
where $P_R(w_i,\dots)$ denotes the output of the reader after reading $w_{i-1}$, and
$P_\mathit{Decoder}(w_i|\dots;h_N)$ is the
output of the decoder at time $i-1$, with $h_N$ being the vector representation created by
the reader network for the entire input sequence.


To implement the Tradeoff Hypothesis, we train NEAT to solve
language modeling and reconstruction with minimal attention, i.e., the
network minimizes the expected loss:
\begin{equation}\label{eq:objective}
  Q(\theta) := \operatorname{E}_{\boldsymbol w, \boldsymbol\omega}\left[L(\boldsymbol{\omega}|\boldsymbol w, \theta) + \alpha \cdot \|\boldsymbol\omega\|_{\ell_1}\right]
\end{equation}
where word sequences $\boldsymbol
w$ are drawn from a corpus, and $\boldsymbol
\omega$ is distributed according to
$P({\boldsymbol\omega}|{\boldsymbol
  w},
\theta)$ as defined in~(\ref{eq:att-prob}).  In (\ref{eq:objective}),
$\|\boldsymbol\omega\|_{\ell_1}$
is the number of words shown to the reader, and $\alpha
> 0$ is a hyperparameter.  The term $\alpha \cdot
\|\boldsymbol\omega\|_{\ell_1}$ encourages NEAT to attend to as few
words as possible.

Note that we make no assumption about linguistic structure -- the only
ingredients of NEAT are the neural architecture, the
objective~(\ref{eq:objective}), and the corpus from which the
sequences $\boldsymbol w$ are drawn.

\subsection{Training}


We follow previous approaches to hard attention in using a combination
of gradient descent and reinforcement learning, and separate the
training of the recurrent networks from the training of $A$.  To train
the reader $R$ and the decoder, we temporarily remove the
attention network $A$, set $\boldsymbol\omega \sim
\mathrm{Binom}(n,p)$ ($n$ sequence length, $p$ a hyperparameter), and
minimize
$\operatorname{E}[L(\boldsymbol{w}|\theta,\boldsymbol\omega)]$ using
stochastic gradient descent, sampling a sequence $\boldsymbol\omega$
for each input sequence.  In effect, NEAT is trained to perform
reconstruction
and language modeling when there is noise in the input.
After $R$ and the decoder have been trained, we fix their
parameters and train $A$ using the REINFORCE rule
\citep{williams_simple_1992}, which performs stochastic gradient
descent using the estimate
\begin{equation}
   \frac{1}{|B|} \sum_{\boldsymbol{w} \in
    B;\boldsymbol\omega} \left( L(\boldsymbol \omega|\boldsymbol w, \theta) + \alpha \cdot \|\boldsymbol\omega\|_{\ell_1}\right) \partial_{\theta_A} \left(\log
    P(\boldsymbol\omega|\boldsymbol{w}, \theta)\right)
\end{equation}
for the gradient $\partial_{\theta_A} Q$. Here, $B$ is a minibatch, $\boldsymbol\omega$ is sampled from
$P(\boldsymbol\omega|\boldsymbol{w}, \theta)$, and $\theta_A \subset
\theta$ is the set of parameters of $A$.
For reducing the variance of
this estimator, we subtract in the $i$-th step an estimate of the
expected loss:
\begin{equation}
\begin{split}
U(\boldsymbol w, \boldsymbol \omega_{1 \dots i-1}) :=
\operatorname{E}_{\boldsymbol\omega_{i \dots N}}[
  L(\boldsymbol\omega_{1 \dots i-1} \boldsymbol\omega_{i \dots N}|{\boldsymbol w}, \theta) \\ 
  +\ \alpha \cdot \|\boldsymbol\omega\|_{\ell_1}]  
\end{split}
\end{equation}
We compute the expected loss using an LSTM
that we train simultaneously with $A$ to predict $L + \alpha \|\boldsymbol\omega\|_{\ell_1}$ based on $\boldsymbol w$ and $\boldsymbol\omega_{1\dots i-1}$.  To make learning more stable, we add
an entropy term encouraging the distribution to be smooth, following
\cite{xu_show_2015}.  The parameter updates to $A$ are thus:
\begin{equation}
\begin{split}
  \sum_{\boldsymbol{w}, \boldsymbol\omega} \sum_i
  \left(L(\boldsymbol{\omega}|\boldsymbol{w}, \theta) + \alpha
    \|\boldsymbol\omega\|_{\ell_1} - U(\boldsymbol w, \boldsymbol\omega_{1 \dots i-1}) \right)\\
  \cdot\ \partial_{\theta_A} \left(log\ P(\omega_i|\boldsymbol\omega_{1 \dots i-1}, \boldsymbol{w}, \theta)\right) \\
  - \gamma\ \partial_{\theta_A} \left(\sum_{\boldsymbol{w},
      \boldsymbol\omega} \sum_i
    \operatorname{H}[P(\omega_i|\boldsymbol\omega_{1,\dots,i-1},\boldsymbol{w},\theta)] \right)
\end{split}
\end{equation}
where $\gamma$ is a hyperparameter, and $\operatorname{H}$ the entropy.




\section{Methods}
\label{sec:methods}


Our aim is to evaluate how well NEAT predicts human fixation behavior
and reading times. Furthermore, we want show that known
qualitative properties emerge from the Tradeoff Hypothesis, even
though no prior knowledge about useful features is hard-wired in
NEAT.


\subsection{Training Setup}

For both the reader and the decoder networks, we choose a one-layer
LSTM network with 1,000 memory cells.  The attention network is a
one-layer feedforward network.  For the loss estimator $U$, we use a
bidirectional LSTM with 20 memory cells.  Input data is split into
sequences of 50 tokens, which are used as the input sequences for
NEAT, disregarding sentence boundaries.  Word embeddings have 100
dimensions, are shared between the reader and the attention network,
and are only trained during the training of the reader.  The
vocabulary consists of the 10,000 most frequent words from the
training corpus.  We trained NEAT on the training set of the
\emph{Daily Mail} section of the corpus described by
\citet{hermann_teaching_2015}, which consists of 195,462 articles from
the \emph{Daily Mail} newspaper, containing approximately 200 million
tokens.  The recurrent networks and the attention network were each
trained for one epoch.  For initialization, weights are drawn from the
uniform distribution.  We set $\alpha = 5.0$, $\gamma = 5.0$, and used
a constant learning rate of $0.01$ for $A$.

\subsection{Corpus}

For evaluation, we use the English section of the Dundee corpus
\citep{kennedy_parafoveal--foveal_2005}, which consists of 20~texts
from \emph{The Independent}, annotated with eye-movement data from ten
English native speakers. Each native speakers read all 20~texts and
answered a comprehension question after each text.  We split the
Dundee corpus into a development and a test set, with texts 1--3
constituting the development set.  The development set consists of
78,300 tokens, and the test set of 281,911 tokens.
For evaluation, we removed the datapoints removed by
\cite{demberg_data_2008}, mainly consisting of words at the beginning
or end of lines, outliers, and cases of track loss.  Furthermore, we
removed datapoints where the word was outside of the vocabulary of the
model, and those datapoints mapped to positions 1--3 or 48--50 of a
sequence when splitting the data. After preprocessing, 62.9\% of the
development tokens and 64.7\% of the test tokens remained.  To obtain
the number of fixations on a token and reading times, we used the
eye-tracking measures computed by \cite{demberg_data_2008}.  The
overall fixation rate was 62.1\% on the development set, and 61.3\% on
the test set.



The development set was used to run preliminary versions of the human evaluation
studies, and to determine the human skipping rate (see Section~\ref{sec:results}).
All the results reported in this paper were computed
on the test set, which remained unseen until the model was final.


\section{Results and Discussion}
\label{sec:results}



Throughout this section, we consider the following baselines for the
attention network: \emph{random attention} is defined by
$\boldsymbol\omega \sim \mathrm{Binom}(n,p)$, with $p = 0.62$, the
human fixation rate in the development set.  For \emph{full
  attention}, we take $\boldsymbol\omega = 1$, i.e., all words are
fixated.  We also derive fixation predictions from \emph{full
  surprisal}, \emph{word frequency}, and \emph{word length} by
choosing a threshold such that the resulting fixation rate matches the
human fixation rate on the development set.


\subsection{Quantitative Properties}

\begin{figure*}[tb]
{\tiny\input{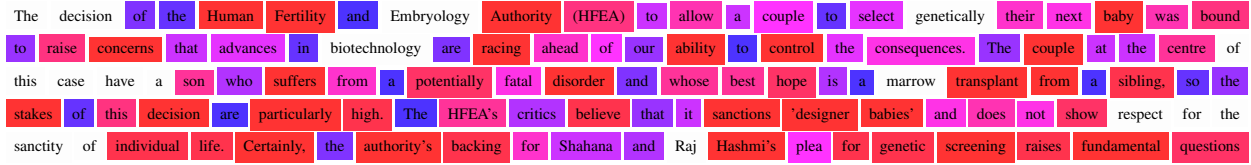}}

\vspace{2ex}

{\tiny\input{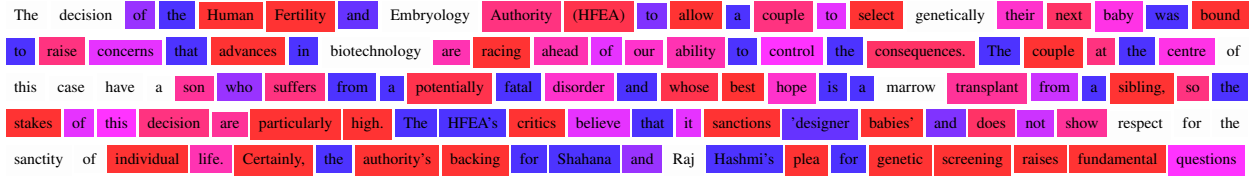}}
\caption{Top: Heatmap showing human fixation probabilities, as
  estimated from the ten readers in the Dundee corpus. In cases of
  track loss, we replaced the missing value with the corresponding
  reader's overall fixation rate. Bottom: Heatmap showing fixation
  probabilities simulated by NEAT. Color gradient ranges from blue
  (low probability) to red (high probability); words without color are
  at the beginning or end of a sequence, or out of vocabulary.}
\label{fig:heatmaps}
\end{figure*}

By averaging over all possible fixation sequences, NEAT defines for
each word in a sequence a probability that it will be fixated.  This
probability is not efficiently computable, so we approximate it by
sampling a sequence $\boldsymbol\omega$ and taking the probabilities
$P(\omega_{i} = 1| \omega_{1 \dots i-1}, \boldsymbol{w})$ for $i = 1, \dots,
50$.  These simulated fixation probabilities can be interpreted as
defining a distribution of attention over the input sequence.
Figure~\ref{fig:heatmaps} shows heatmaps of the simulated and human
fixation probabilities, respectively, for the beginning of a text from
the Dundee corpus.  While some differences between simulated and human
fixation probabilities can be noticed, there are similarities in the
general qualitative features of the two heatmaps.  In particular,
function words and short words are less likely to be fixated than
content words and longer words in both the simulated and the human
data.

\paragraph{Reconstruction and Language Modeling}

We first evaluate NEAT intrinsically by measuring how successful the
network is at predicting the next word and reconstructing the input
while minimizing the number of fixations.  We compare perplexity on
reconstruction and language modeling for
$\boldsymbol\omega \sim P(\boldsymbol\omega | \boldsymbol w, \theta)$.
In addition to the baselines, we run NEAT on the fixations generated
by the human readers of the Dundee corpus, i.e., we use the human
fixation sequence as $\boldsymbol \omega$ instead of the fixation
sequence generated by $A$ to compute perplexity. This will tell us to
what extent the human behavior minimizes the NEAT
objective~(\ref{eq:objective}).

The results are given in Table~\ref{fig:intrinsic}.  In all settings,
the fixation rates are similar ($60.4\%$ to $62.1\%$)
which makes the perplexity figures directly comparable.  While NEAT
has a higher perplexity on both tasks compared to full attention, it
considerably outperforms random attention. It also outperforms the
word length, word frequency, and full surprisal baselines.
The perplexity on human fixation sequences is similar to that achieved
using word frequency. Based on these results, we conclude that
REINFORCE successfully optimizes the objective~(\ref{eq:objective}).


\begin{table*}[tb]
\begin{center}
\begin{tabular}{l@{\hspace{1ex}~}l@{\hspace{1ex}~}l@{\hspace{1ex}~}l@{\hspace{1ex}~}l@{\hspace{1ex}~}l@{\hspace{1ex}~}l@{\hspace{1ex}~}l}
\hline
                  &   NEAT & Rand. Att. & Word Len. & Word Freq.& Full Surp. & Human  & Full Att. \\ \hline
Language Modeling &  180        & 333            & 230   & 219 & 211 & 218/170 &   107 \\
Reconstruction    & 4.5      &  56  &    40     & 39   &  34 & 39/31 &  1.6 \\
Fixation Rate     & 60.4\%   & 62.1\%                &   62.1\% &   62.1\% & 62.1\% & 61.3\%/72.0\% & 100\%\\
\hline
\end{tabular}
\end{center}

\caption{Performance on language modeling and reconstruction as measured
  by perplexity. Random attention is an upper bound on perplexity,
  while full attention is a lower bound. For the human baseline, we
  give two figures, which differ in the treatment of missing data. The
  first figure is obtained when replacing missing values with a random 
  variable $\omega \sim \mathrm{Binom}(n,0.61)$; the second results
  from replacing missing values with $1$.
}\label{fig:intrinsic}
\end{table*}

\paragraph{Likelihood of Fixation Data}

Human reading behavior is stochastic in the sense that different runs
of eye-tracking experiments such as the ones recorded in the Dundee
corpus yield different eye-movement sequences.  NEAT is also
stochastic, in the sense that, given a word sequence ${\bf w}$, it
defines a probability distribution over fixation sequences
{$\boldsymbol\omega$}.  Ideally, this distribution should be close to
the actual distribution of fixation sequences produced by humans
reading the sequence, as measured by perplexity.

We find that the perplexity of the fixation sequences produced by the
ten readers in the Dundee corpus under NEAT is 1.84.  A perplexity of
2.0 corresponds to the random baseline $\mathrm{Binom}(n,0.5)$, and a
perplexity of 1.96 to random attention $\mathrm{Binom}(n,0.62)$.
As a lower bound on what can achieved with models disregarding the
context, using the human fixation rates for each word as
probabilities, we obtain a perplexity of 1.68.

\begin{table}[tb]
\begin{center}
\begin{tabular}{l@{\hspace{1ex}}lll}
\hline
                   & Acc  & F1$_{\mathrm{fix}}$ & F1$_{\mathrm{skip}}$ \\ \hline
NEAT & 63.7 & 70.4 & 53.0 \\ \hline 
\multicolumn{4}{c}{Supervised Models} \\ \hline
{\small\cite{nilsson_learning_2009}}
& 69.5 & 75.2 & 62.6 \\   
{\small\cite{matthies_blinkers_2013}} 
&  69.9 & 72.3 & 66.1 \\ \hline  
\multicolumn{4}{c}{Human Performance and Baselines} \\ \hline
Random Baseline & 52.6 & 62.1 & 37.9 \\ 
Full Surprisal & 64.1 & 70.7 & 53.6 \\ 
Word Frequency & 67.9 & 74.0 & 58.3 \\
Word Length    & 68.4 & 77.1 & 49.0 \\
Human & 69.5 & 76.6 & 53.6 \\ \hline
\end{tabular}
\end{center}
\caption{Evaluation of fixation sequence predictions against human
  data. For the human baseline, we predicted the $n$-th reader's fixations
  by taking the fixations of the $n+1$-th reader (with missing
  values replaced by reader average), averaging the resulting scores
  over the ten readers.
}\label{fig:disc-eval}
\end{table}

\paragraph{Accuracy of Fixation Sequences}
\label{sec:fixation_seq_accuracy}

Previous work on supervised models for modeling fixations
\citep{nilsson_learning_2009,matthies_blinkers_2013} has been
evaluated by measuring the overlap of the fixation sequences produced
by the models with those in the Dundee corpus.  For NEAT, this method
of evaluation is problematic as differences between model predictions
and human data may be due to differences in the rate of skipping, and
due to the inherently stochastic nature of fixations.  We therefore
derive model predictions by rescaling the simulated fixation
probabilities so that their average equals the fixation rate in the
development set, and then greedily take the maximum-likelihood
sequence.  That is, we predict a fixation if the rescaled probability
is greater than $0.5$, and a skip otherwise.  As in previous work, we
report the accuracy of fixations and skips, and also separate F1
scores for fixations and skips.  As lower and upper bounds, we use the
random baseline $\boldsymbol\omega \sim \mathrm{Binom}(n, 0.62)$ and
the agreement of the ten human readers, respectively.  The results are
shown in Table~\ref{fig:disc-eval}.  NEAT clearly outperforms the
random baseline and shows results close to full surprisal (where we
apply the same rescaling and thresholding as for NEAT). This is
remarkable given that NEAT has access to only $60.4\%$ of the words in
the corpus in order to predict skipping, while full surprisal has
access to all the words.

Word frequency and word length perform well, almost reaching the
performance of supervised models.  This shows that the bulk of
skipping behavior is already explained by word frequency and word
length effects. Note, however, that NEAT is completely unsupervised,
and does not know that it has to pay attention to word frequency; this
is something the model is able to infer.


\paragraph{Restricted Surprisal and Reading Times}


To evaluate the predictions NEAT makes for reading times, we use
linear mixed-effects models containing restricted surprisal derived
from NEAT for the Dundee test set. The mixed models also include a set
of standard baseline predictors, viz., word length, log word
frequency, log frequency of the previous word, launch distance,
landing position, and the position of the word in the sentence.  We
treat participants and items as random factors.  As the dependent
variable, we take first pass duration, which is the sum of the
durations of all fixations from first entering the word to first
leaving it.  We compare against \emph{full surprisal} as an upper
bound and against \emph{random surprisal} as a lower bound. Random
surprisal is surprisal computed by a model with random attention; this
allows us to assess how much surprisal degrades when only 60.4\% of
all words are fixated, but no information is available as to which
words should be fixated. The results in Table~\ref{fig:mixed} show
that restricted surprisal as computed by NEAT, full surprisal, and
random surprisal are all significant predictors of reading time.

In order to compare the three surprisal estimates, we therefore need a
measure of effect size. For this, we compare the model fit of the
three mixed effects models using deviance, which is defined as the
difference between the log likelihood of the model under consideration
minus the log likelihood of the baseline model, multiplied by $-2$.
Higher deviance indicates greater improvement in model fit over the
baseline model. We find that the mixed model that includes restricted
surprisal achieves a deviance of~867, compared to the model containing
only the baseline features. With full surprisal, we obtain a deviance
of~980. On the other hand, the model including random surprisal
achieves a lower deviance of~832.

This shows that restricted surprisal as computed by NEAT not only
significantly predicts reading times, it also provides an improvement
in model fit compared to the baseline predictors. Such an improvement
is also observed with random surprisal, but restricted surprisal
achieves a greater improvement in model fit. Full surprisal achieves an
even greater improvement, but this is not unexpected, as full
surprisal has access to all words, unlike NEAT or random surprisal,
which only have access to 60.4$\%$ of the words.

\begin{table}[tb]
\begin{center}
\begin{tabular}{l@{}rr@{~~}r@{}l}
\hline
                    &   $\beta$ & SE & $t$     \\ \hline
(Intercept)         &  247.43   &  7.14     & 34.68&* \\
Word Length         &   12.92   &  0.21     & 60.62&*  \\
Previous Word Freq. & $-$5.28   &  0.28  & $-$18.34&* \\
Prev. Word Fixated  &$-$24.67   &  0.81  & $-$30.55&* \\
Launch Distance     & $$-0.01   &  0.01  &  $-$0.37&   \\
Obj. Landing Pos.   & $-$8.07   &  0.20  & $-$41.25&* \\
Word Pos. in Sent.  & $-$0.10   &  0.03  &  $-$2.98&*   \\
Log Word Freq.      & $-$1.59   &  0.21  &  $-$7.73&*  \\ \hline 
Resid. Random Surprisal& 2.69   &  0.10  &    29.27&* \\
Resid. Restr. Surprisal& 2.75   &  0.12  &    23.66&*     \\
Resid. Full Surprisal  & 2.99   &  0.12  &    25.23&*      \\ \hline
\end{tabular}
\end{center}

\caption{Linear mixed effects models for first pass duration. The
  first part of the table shows the coefficients, standard errors, and $t$ values for
  the predictors in the baseline model. The second part of the table gives
  the corresponding values for random surprisal, restricted surprisal
  computed by NEAT, and full surprisal, residualized against the baseline predictors, in 
  three models obtained by adding these predictors.}\label{fig:mixed}
\end{table}

\subsection{Qualitative Properties}

We now examine the second key question we defined in
Section~\ref{sec:intro}, investigating the qualitative features of the
simulated fixation sequences.  We will focus on comparing the
predictions of NEAT with that of word frequency, which performs
comparably at the task of predicting fixation sequences (see
Section~\ref{sec:fixation_seq_accuracy}). We show NEAT nevertheless
makes relevant predictions that go beyond frequency.

\paragraph{Fixations of Successive Words}

While predictors derived from word frequency treat the decision
whether to fixate or skip words as independent, humans are more likely
to fixate a word when the previous word was
skipped~\citep{rayner_eye_1998}.  This effect is also seen in NEAT.
More precisely, both in the human data and in the simulated fixation
data, the conditional fixation probability $P(\omega_i =
1|\omega_{i-1} = 1)$ is lower than the marginal probability $P(\omega_i =
1)$.  The ratio of these probabilities is $0.85$ in the human data,
and $0.81$ in NEAT.  The threshold predictor derived from word
frequency also shows this effect (as the frequencies of successive
words are not independent), but it is weaker (ratio $0.91$).

To further test the context dependence of NEAT's fixation behavior, we
ran a mixed model predicting the fixation probabilities simulated by
NEAT, with items as random factor and the log frequency of word $i$ as
predictor. Adding $\omega_{i-1}$ as a predictor results in a
significant improvement in model fit (deviance = 4,798, $t = 71.3$).
This shows that NEAT captures the context dependence of fixation
sequences to an extend that goes beyond word frequency alone.

\begin{table}[tb]
\begin{center}
\begin{tabular}{llllllllll}
\hline
          & Human   &     NEAT  & Word Freq.   \\ \hline
ADJ       & 78.9   (2)    &  72.8 (1)         & 98.4    (3)       \\
ADP       & 46.1    (8)   &  53.8 (8)         &  21.6     (9)       \\ 
ADV       & 70.4 (3)      &  67.2 (4)         &   96.4   (4)        \\
CONJ      & 36.7    (11)  &   50.7 (9)        &   14.6    (10)       \\
DET       & 45.2    (9)   &  44.8 (11)        &    22.9     (8)     \\
NOUN      & 80.3   (1)    &  69.8 (2)         &   98.7   (2)        \\
NUM       & 63.3   (6)    &  71.5 (3)         &    99.5    (1)      \\
PRON      & 49.2    (7)   &   57.0 (7)        &   42.6    (7)       \\
PRT       & 37.4    (10)  &   46.7 (10)       &   13.9   (11)         \\
VERB      & 66.7   (5)    &   64.7 (5)        &     74.4  (5)       \\
X         & 68.6  (4)     &   67.8 (3)        &     69.0   (6)      \\ \hline
Spearman's $\rho$ &       &   0.85            &     0.84         \\
Pearson's $r$     &       &   0.92            &    0.94          \\
MSE       &               &   57              & 450              \\
\hline
\end{tabular}
\end{center}

\caption{Actual and simulated fixation probabilities (in $\%$) by PoS
  tag, with the ranks given in brackets, and correlations and mean
  squared error relative to human data.}\label{fig:pos}
\end{table}


\paragraph{Parts of Speech}

Part of speech categories are known to be a predictor of fixation
probabilities, with content words being more likely to be fixated than
function words \citep{carpenter_what_1983}.  In Table~\ref{fig:pos},
we give the simulated fixation probabilities and the human fixation
probabilities estimated from the Dundee corpus for the tags of the
Universal PoS tagset \citep{petrov_universal_2012}, using the PoS
annotation of~\cite{barrett_dundee_2015}. We again compare with the
probabilities of a threshold predictor derived from word
frequency.\footnote{We omit the tag ``.'' for punctuation, as
  punctuation characters are not treated as separate tokens in
  Dundee.}
NEAT captures the differences between PoS categories well, as
evidenced by the high correlation coefficients.  The content word
categories ADJ, ADV, NOUN, VERB and X consistently show higher
probabilities than the function word categories.  While the
correlation coefficients for word frequency are very similar, the
numerical values of the simulated probabilities are closer to the
human ones than those derived from word frequency, which tend towards
more extreme values. This difference can be seen clearly if we compare
the mean squared error, rather than the correlation, with the human
fixation probabilities (last row of Table~\ref{fig:pos}).












\paragraph{Correlations with Known Predictors}

In the literature, it has been observed that skipping correlates with
predictability (surprisal), word frequency, and word length
\citep[p.\,387]{rayner_eye_1998}. These correlations are also observed
in the human skipping data derived from Dundee, as shown in
Table~\ref{fig:correlations}. (Human fixation probabilities were
obtained by averaging over the ten readers in Dundee.)

Comparing the known predictors of skipping with NEAT's simulated
fixation probabilities, similar correlations as in the human data are
observed. We observe that the correlations with surprisal are stronger
in NEAT, considering both restricted surprisal and full surprisal as
measures of predictability.


\begin{table}[tb]
\begin{center}
\begin{tabular}{lrr}
\hline              
                    &  Human & NEAT \\
\hline
Restricted Surprisal& 0.465 & 0.762 \\
Full Surprisal      & 0.512 & 0.720 \\
Log Word Freq.      & $-$0.608 & $-$0.760 \\
Word Length         & 0.663 & 0.521 \\
\hline
\end{tabular}
\end{center}
\caption{Correlations between human and NEAT fixation probabilities and
  known predictors}\label{fig:correlations}
\end{table}

\section{Conclusions}
\label{sec:concl}

We investigated the hypothesis that human reading strategies optimize
a tradeoff between precision of language understanding and economy of
attention.  We made this idea explicit in NEAT, a neural reading
architecture with hard attention that can be trained end-to-end to
optimize this tradeoff.  Experiments on the Dundee corpus show that
NEAT provides accurate predictions for human skipping behavior. It
also predicts reading times
, even though
it only has access to 60.4\% of the words in the corpus in order to
estimate surprisal. Finally, we found that known qualitative
properties of skipping emerge in our model, even though they were not
explicitly included in the architecture, such as context dependence of
fixations, differential skipping rates across parts of speech, and
correlations with other known predictors of human reading behavior.





\bibliographystyle{acl}
\bibliography{references}

\begin{thebibliography}{35}
\providecommand{\natexlab}[1]{#1}

\bibitem[{Ba et~al.(2015)Ba, Salakhutdinov, Grosse, and
  Frey}]{ba_learning_2015}
Ba, Jimmy, Ruslan~R. Salakhutdinov, Roger~B. Grosse, and Brendan~J. Frey. 2015.
\newblock Learning wake-sleep recurrent attention models.
\newblock In {\em Advances in Neural Information Processing Systems\/}. pages
  2575--2583.

\bibitem[{Bahdanau et~al.(2015)Bahdanau, Cho, and
  Bengio}]{bahdanau_neural_2015}
Bahdanau, Dzmitry, Kyunghyun Cho, and Yoshua Bengio. 2015.
\newblock Neural machine translation by jointly learning to align and
  translate.
\newblock In {\em Proceedings of the International Conference on Learning
  Representations\/}.

\bibitem[{Barrett et~al.(2015)Barrett, Agi{\' c}, and
  S{\o}gaard}]{barrett_dundee_2015}
Barrett, Maria, {\v Z}eljko Agi{\' c}, and Anders S{\o}gaard. 2015.
\newblock The {Dundee} treebank.
\newblock In {\em Proceedings of the 14th International Workshop on Treebanks
  and Linguistic Theories\/}. pages 242--248.

\bibitem[{Bicknell and Levy(2010)}]{bicknell_rational_2010-1}
Bicknell, Klinton and Roger Levy. 2010.
\newblock A rational model of eye movement control in reading.
\newblock In {\em Proceedings of the 48th Annual Meeting of the Association for
  Computational Linguistics\/}. pages 1168--1178.

\bibitem[{Carpenter and Just(1983)}]{carpenter_what_1983}
Carpenter, P.~A. and M.~A. Just. 1983.
\newblock What your eyes do while your mind is reading.
\newblock In K.~Rayner, editor, {\em Eye Movements in Reading\/}, Academic
  Press, New York, pages 275--307.

\bibitem[{Demberg and Keller(2008)}]{demberg_data_2008}
Demberg, Vera and Frank Keller. 2008.
\newblock Data from eye-tracking corpora as evidence for theories of syntactic
  processing complexity.
\newblock {\em Cognition\/} 109(2):193--210.

\bibitem[{Engbert et~al.(2002)Engbert, Longtin, and
  Kliegl}]{engbert_dynamical_2002}
Engbert, Ralf, André Longtin, and Reinhold Kliegl. 2002.
\newblock A dynamical model of saccade generation in reading based on spatially
  distributed lexical processing.
\newblock {\em Vision Research\/} 42(5):621--636.

\bibitem[{Engbert et~al.(2005)Engbert, Nuthmann, Richter, and
  Kliegl}]{engbert_swift:_2005}
Engbert, Ralf, Antje Nuthmann, Eike~M. Richter, and Reinhold Kliegl. 2005.
\newblock {SWIFT}: A dynamical model of saccade generation during reading.
\newblock {\em Psychological Review\/} 112(4):777--813.

\bibitem[{Frank and Bod(2011)}]{frank:bod:11}
Frank, S.L. and R.~Bod. 2011.
\newblock Insensitivity of the human sentence-processing system to hierarchical
  structure.
\newblock {\em Psychological Science\/} 22:829--834.

\bibitem[{Hale(2001)}]{hale_probabilistic_2001}
Hale, John. 2001.
\newblock A probabilistic {Earley} parser as a psycholinguistic model.
\newblock In {\em Proceedings of Conference of the North American Chapter of
  the Association for Computational Linguistics\/}. volume~2, pages 159--166.

\bibitem[{Hara et~al.(2012)Hara, Kano, and Aizawa}]{hara_predicting_2012}
Hara, Tadayoshi, Daichi Mochihashi~Yoshinobu Kano, and Akiko Aizawa. 2012.
\newblock Predicting word fixations in text with a {CRF} model for capturing
  general reading strategies among readers.
\newblock In {\em Proceedings of the 1st Workshop on Eye-tracking and Natural
  Language Processing\/}. pages 55--70.

\bibitem[{Henderson(2003)}]{Henderson:03}
Henderson, John. 2003.
\newblock Human gaze control in real-world scene perception.
\newblock {\em Trends in Cognitive Sciences\/} 7:498--504.

\bibitem[{Hermann et~al.(2015)Hermann, Ko{\v c}isk{\`y}, Grefenstette,
  Espeholt, Kay, Suleyman, and Blunsom}]{hermann_teaching_2015}
Hermann, Karl~Moritz, Tom{\'a}{\v s} Ko{\v c}isk{\`y}, Edward Grefenstette,
  Lasse Espeholt, Will Kay, Mustafa Suleyman, and Phil Blunsom. 2015.
\newblock Teaching machines to read and comprehend.
\newblock ArXiv:1506.03340.

\bibitem[{Hochreiter and Schmidhuber(1997)}]{hochreiter_long_1997}
Hochreiter, Sepp and J{\"u}rgen Schmidhuber. 1997.
\newblock Long short-term memory.
\newblock {\em Neural Computation\/} 9(8):1735--1780.

\bibitem[{Kennedy and Pynte(2005)}]{kennedy_parafoveal--foveal_2005}
Kennedy, Alan and Joël Pynte. 2005.
\newblock Parafoveal-on-foveal effects in normal reading.
\newblock {\em Vision Research\/} 45(2):153--168.

\bibitem[{Levy(2008)}]{levy_expectation-based_2008}
Levy, Roger. 2008.
\newblock Expectation-based syntactic comprehension.
\newblock {\em Cognition\/} 106(3):1126--1177.

\bibitem[{Matthies and Søgaard(2013)}]{matthies_blinkers_2013}
Matthies, Franz and Anders Søgaard. 2013.
\newblock With blinkers on: Robust prediction of eye movements across readers.
\newblock In {\em Proceedings of the Conference on Empirical Methods in Natural
  Language Processing\/}. pages 803--807.

\bibitem[{McDonald and Shillcock(2003{\natexlab{a}})}]{mcdonald_eye_2003}
McDonald, Scott~A. and Richard~C. Shillcock. 2003{\natexlab{a}}.
\newblock Eye movements reveal the on-line computation of lexical probabilities
  during reading.
\newblock {\em Psychological Science\/} 14(6):648--652.

\bibitem[{McDonald and Shillcock(2003{\natexlab{b}})}]{mcdonald_low-level_2003}
McDonald, Scott~A. and Richard~C. Shillcock. 2003{\natexlab{b}}.
\newblock Low-level predictive inference in reading: the influence of
  transitional probabilities on eye movements.
\newblock {\em Vision Research\/} 43(16):1735--1751.

\bibitem[{Mikolov et~al.(2010)Mikolov, Karafi{\'a}t, Burget, {\v C}ernock{\'y},
  and Khudanpur}]{mikolov_recurrent_2010}
Mikolov, Tom{\'a}{\v s}, Martin Karafi{\'a}t, Luk{\'a}{\v s} Burget, Jan {\v
  C}ernock{\'y}, and Sanjeev Khudanpur. 2010.
\newblock Recurrent neural network based language model.
\newblock In {\em Proceedings of Interspeech\/}. pages 1045--1048.

\bibitem[{Mnih et~al.(2014)Mnih, Heess, Graves, and
  {others}}]{mnih_recurrent_2014}
Mnih, Volodymyr, Nicolas Heess, Alex Graves, and {others}. 2014.
\newblock Recurrent models of visual attention.
\newblock In {\em Advances in Neural Information Processing Systems\/}. pages
  2204--2212.

\bibitem[{Nilsson and Nivre(2009)}]{nilsson_learning_2009}
Nilsson, Mattias and Joakim Nivre. 2009.
\newblock Learning where to look: Modeling eye movements in reading.
\newblock In {\em Proceedings of the 13th Conference on Computational Natural
  Language Learning\/}. pages 93--101.

\bibitem[{Nilsson and Nivre(2010)}]{nilsson_towards_2010}
Nilsson, Mattias and Joakim Nivre. 2010.
\newblock Towards a data-driven model of eye movement control in reading.
\newblock In {\em Proceedings of the Workshop on Cognitive Modeling and
  Computational Linguistics\/}. pages 63--71.

\bibitem[{Petrov et~al.(2012)Petrov, Das, and McDonald}]{petrov_universal_2012}
Petrov, Slav, Dipanjan Das, and Ryan~T. McDonald. 2012.
\newblock A universal part-of-speech tagset.
\newblock In {\em Proceedings of the 8th International Conference on Language
  Resources and Evaluation\/}. pages 2089--2096.

\bibitem[{Rayner(1998)}]{rayner_eye_1998}
Rayner, K. 1998.
\newblock Eye movements in reading and information processing: 20 years of
  research.
\newblock {\em Psychological Bulletin\/} 124(3):372--422.

\bibitem[{Rayner(2009)}]{rayner_eye_2009}
Rayner, Keith. 2009.
\newblock Eye movements in reading: Models and data.
\newblock {\em Journal of Eye Movement Research\/} 2(5):1--10.

\bibitem[{Rayner and Reichle(2010)}]{rayner_models_2010-1}
Rayner, Keith and Erik~D. Reichle. 2010.
\newblock Models of the reading process.
\newblock {\em Wiley Interdisciplinary Reviews: Cognitive Science\/}
  1(6):787--799.

\bibitem[{Reichle et~al.(1998)Reichle, Pollatsek, Fisher, and
  Rayner}]{reichle_toward_1998}
Reichle, E.~D., A.~Pollatsek, D.~L. Fisher, and K.~Rayner. 1998.
\newblock Toward a model of eye movement control in reading.
\newblock {\em Psychological Review\/} 105(1):125--157.

\bibitem[{Reichle et~al.(2009)Reichle, Warren, and
  McConnell}]{reichle_using_2009}
Reichle, E.~D., T.~Warren, and K.~McConnell. 2009.
\newblock Using {EZ Reader} to model the effects of higher level language
  processing on eye movements during reading.
\newblock {\em Psychonomic Bulletin \& Review\/} 16(1):1--21.

\bibitem[{Reichle et~al.(2003)Reichle, Rayner, and Pollatsek}]{reichle_ez_2003}
Reichle, Erik~D., Keith Rayner, and Alexander Pollatsek. 2003.
\newblock The {EZ Reader} model of eye-movement control in reading: Comparisons
  to other models.
\newblock {\em Behavioral and Brain Sciences\/} 26(04):445--476.

\bibitem[{Smith and Levy(2013)}]{smith_effect_2013}
Smith, Nathaniel~J. and Roger Levy. 2013.
\newblock The effect of word predictability on reading time is logarithmic.
\newblock {\em Cognition\/} 128(3):302--319.

\bibitem[{Sutskever et~al.(2014)Sutskever, Vinyals, and
  Le}]{sutskever_sequence_2014}
Sutskever, Ilya, Oriol Vinyals, and Quoc~VV Le. 2014.
\newblock Sequence to sequence learning with neural networks.
\newblock In {\em Advances in Neural Information Processing Systems\/}. pages
  3104--3112.

\bibitem[{Van~Gompel and Pickering(2007)}]{van_gompel_syntactic_2007}
Van~Gompel, Roger~PG and Martin~J. Pickering. 2007.
\newblock Syntactic parsing.
\newblock In {\em The Oxford Handbook of Psycholinguistics\/}, Oxford
  University Press, pages 289--307.

\bibitem[{Williams(1992)}]{williams_simple_1992}
Williams, Ronald~J. 1992.
\newblock Simple statistical gradient-following algorithms for connectionist
  reinforcement learning.
\newblock {\em Machine Learning\/} 8(3-4):229--256.

\bibitem[{Xu et~al.(2015)Xu, Ba, Kiros, Courville, Salakhutdinov, Zemel, and
  Bengio}]{xu_show_2015}
Xu, Kelvin, Jimmy Ba, Ryan Kiros, Aaron Courville, Ruslan Salakhutdinov,
  Richard Zemel, and Yoshua Bengio. 2015.
\newblock Show, attend and tell: Neural image caption generation with visual
  attention.
\newblock ArXiv:1502.03044.

\end{thebibliography}

\end{document}